# Segmentation of cracks in 3d images of fiber reinforced concrete using deep learning

**Anna Nowacka[1,2], Katja Schladitz[1], Szymon Grzesiak[2], Matthias Pahn[2]**

1. Department of Image Processing, Fraunhofer Institute for Industrial Mathematics, Kaiserslautern, Germany
2. Department of Civil Engineering, University of Kaiserslautern-Landau, Kaiserslautern, Germany

Corresponding author: *anna.nowacka@itwm.fraunhofer.de*

**ABSTRACT**

Cracks in concrete structures are very common and are an integral part of this heterogeneous material. Characteristics of cracks induced by standardized tests yield valuable information about the tested concrete formulation and its mechanical properties. Observing cracks on the surface of the concrete structure leaves a wealth of structural information unused. Computed tomography enables looking into the sample without interfering or destroying the microstructure. The reconstructed tomographic images are 3d images, consisting of voxels whose gray values represent local X-ray absorption. In order to identify voxels belonging to the crack, so to segment the crack structure in the images, appropriate algorithms need to be developed. Convolutional neural networks are known to solve this type of task very well given enough and consistent training data. We adapted a 3d version of the well-known U-Net and trained it on semi-synthetic 3d images of real concrete samples equipped with simulated crack structures. Here, we explain the general approach. Moreover, we show how to teach the network to detect also real crack systems in 3d images of varying types of real concrete, in particular of fiber reinforced concrete.

## 1. Introduction

Engineering constructions frequently employ concrete as a primary building material so this is why it is important to understand and model its behavior [1]. The mechanical properties of concrete are routinely experimentally analyzed by standardized tests [2]. In addition to mechanical tests, image processing can be used to examine the properties of concrete [3]. Computed tomography

(CT) offers the new opportunity to image the inner structure of the sample without destroying it [4] [5].

A CT scan yields a 3d gray value image of the concrete sample. Based on this image, the microstructure of the concrete and eventual crack systems can be visualized and analyzed non-destructively [6] [7]. However, CT images are huge, concrete is a very heterogeneous material, and cracks are hard to capture. Thus, dedicated image processing and analysis algorithms have to be developed.

Examining the inner structure of concrete turns into an even more demanding task when reinforcements are taken into account as they vary widely and change the spatial distribution of pores and aggregates [8]. There are various methods of reinforcing concrete elements ranging from bars to distributed fibers made of various materials like steel or glass as described for example in [9] or [10]. CT imaging helps to understand these materials better. A prerequisite for this is however robust and reproducible analysis of the 3d image data [11] which in turn requires proper segmentation of the structural components to be analyzed.

In this work, we continue our research on segmentation of real cracks in concrete in 3d CT images. We extend the capabilities of the machine learning model described in [12], [13], and [14] by applying it to image data yet unseen and unused. Previously, 3d crack segmentation in CT images of concrete has been explored only in [15], [7], and [16]. In [15], three conventional segmentation methods were assessed by examining manually segmented cracks in CT images. In contrast, in [7], those methods were employed on larger CT images, reaching up to $2.000^3$ voxels. Finally, the limitations, robustness, and reliability of automatic 3d crack segmentation methods were discussed in [16].

In [12] the work from [15] was continued, including quantitative comparison of machine learning and classical methods based on semi-synthetic crack images. The 3d U-Net [17], along with a random forest method used for automatic detection of road cracks in [18], were compared to the classical methods proposed in [15]. These methods were fine-tuned in a controlled setting with predefined crack scales. To this end, artificial crack structures of fixed thickness were simulated using fractional Brownian surfaces as available as MATLAB function [19]. Alternatively, minimum-weight surfaces in 3d Voronoi tessellations can be applied [20]. The discrete crack model structures arising from both models can easily be thickened (dilated) with locally varying strength. For details see [13] or [20]. To generate semi-synthetic CT images with cracks, simulated cracks are combined with CT scans of uncracked concrete. Realistic looking cracks within the images are achieved by equipping them with the gray value distribution observed in pores. In these images, the ground truth is known. The semi-synthetic image data therefore allows for objective quantitative comparison and evaluation of methods.

In our previous study [12], the deep learning model was trained on CT images of normal and high-performance concrete. After applying this method to the crack images in high-performance and normal concrete, satisfactory results were obtained and such example can be observed in **Fig. 1**. In this work, we extend this approach to include concretes enforced by polypropylene fibers (PPFRC) and steel fibers (SFRC). We discuss the applicability, limitations, and robustness of our solution and possible ways to improve it.

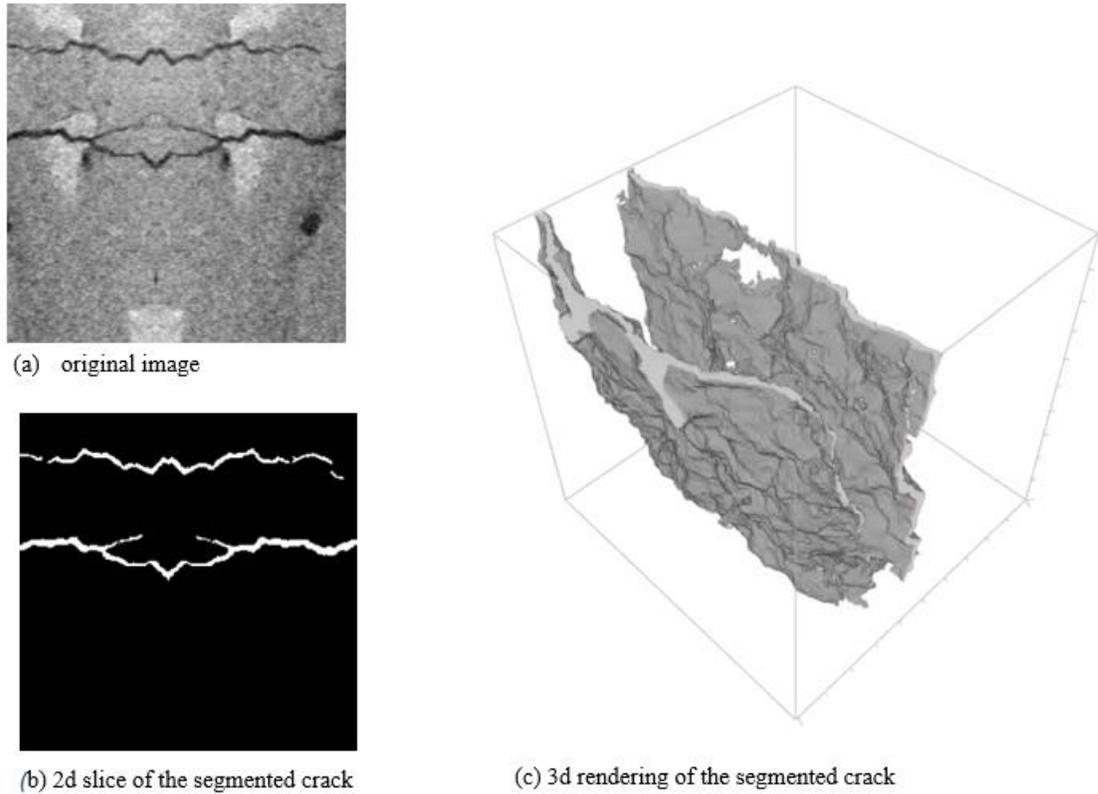

**Fig. 1** Example of the crack segmentation result.

## 2. Materials and methods

In this section we describe sample creation including crack induction, sample characterization by mechanical test, and the CT imaging.

### 2.1. Sample preparation and mechanical test

Samples of concrete were produced using high-performance concrete that contained reinforcement in the form of either steel fibers (SFRC) or polypropylene fibers (PPFRC) [21]. The concrete class is assigned on the basis of compressive strength according to [22]. Based on three cubes, the mean value is 77.2MPa.

The fiber materials differ in their material characteristics and processing method. The stiffness of the steel fibers is 25 times higher than the stiffness of the PP fibers due to the difference in the elastic moduli of the two base materials. Young's modulus of steel is 200GPa while for the PP fibers it is only 8GPa. The heavy steel fibers deposited at the bottom of the sample due to gravity. The specific gravity of PP fibers is 910kg/m³ while for steel it is 7.900kg/m³.

Beam specimens with PP and steel fibers were mechanically tested according to [23] and [24], respectively. The forces were recorded using a load cell and the deflections during the experiments were measured by a displacement transducer. When testing the PPFRC samples, we used crack mouth opening displacement (CMOD) in addition to the regular measurement technique as proposed in [25] and [26], for assessing the deformation of high performance fiber reinforced concrete beams. According to the standard [23], a clip gauge was attached on the bottom surface of the beam specimens to measure the distance between the two opposing edges of the crack as described in [27]. We calculated the stresses $f_L$ in accordance with [28]:

$$f_L = \frac{3 \cdot F \cdot l}{2 \cdot b \cdot h^2} \quad (1)$$

,where $F$ represents the cylinder force (in N); $b$ the width and $h$ the height of the sample; $l$ the distance of the supports.

The parameter $f_L$ was assessed for the four CMOD values 0.5, 1.5, 2.5, and 3.5mm based on the experimental stress-strain curves. The measurement was stopped when the displacement reached 18.7mm. Once this opening is reached, the crack does not close again, which allows further examination of the sample with the induced deformation. From the tested samples, one PPFRC and one SFRC were imaged. The imaged PPFRC specimen is the one with PP fiber dosage 5kg/m³.

For more information on other fiber dosages as well as their influence on the residual bending tensile, splitting tensile, and compressive strength, we refer to [29] and [30].

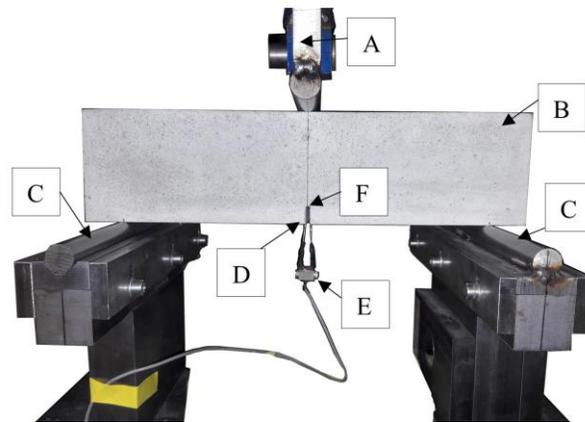

**Fig. 2.** Setup of bending test: A - point of application of symmetrical load from hydraulic jack, B - concrete beam specimen (the dimensions of the beam: 12.91 × 12.05 × 14.61 cm), C supports, D - glued on metal plates to attach the clip gauge, E - clip gauge, F - starter notch (length = 17mm).

## 2.2. CT imaging of the PPFRC sample using a linear accelerator

The PPFRC sample with notch was scanned using a special CT device at Fraunhofer EZRT in Fürth, Germany, featuring a linear accelerator (linac) with a maximum X-ray energy of 8.3MeV and a flat panel detector with a sensitive area of approximately 36×36cm$^2$ and a pixel size of 200μm. The detector was mounted at a distance of 4.5m from the focal spot of the linac at the height of the concrete sample. The concrete sample was placed in a vertical position on a wooden platform that rotated between the linac and the detector. A magnification factor of 1.104 was selected for the measurement on the detector. This results in an effective voxel size of 60.4μm within the object. To create the 3d reconstruction, the sample was rotated a full 360 degrees and 2.400 angles were captured, with each angle exposed for 1 second. The entire measurement process took 40 minutes. **Fig. 3** shows the test setup as well as detector and linac.

## 2.3. CT imaging of the SFRC sample using a laboratory device

The SFRC sample and the test it underwent are described in [31] and [32], respectively. In [31] the details on the sample which is made of normal concrete and reinforced by 60mm long and 1mm thick steel fibers were provided. The amount of fibers is estimated to be 60kg/m$^3$. In [32], the 4-point bending test, which induced the crack system, is explained: two pins support the beam and the force is applied through two loading pins. The sample without notch was scanned using the CT device at Fraunhofer Institute for Industrial Mathematics (ITWM) Kaiserslautern, Germany. This custom made device features a Feinfocus FXE 225 X-ray emitting tube and a PerkinElmer detector with 2.048 × 2.048 pixels. The exemplary setup can be observed in **Fig. 4**.

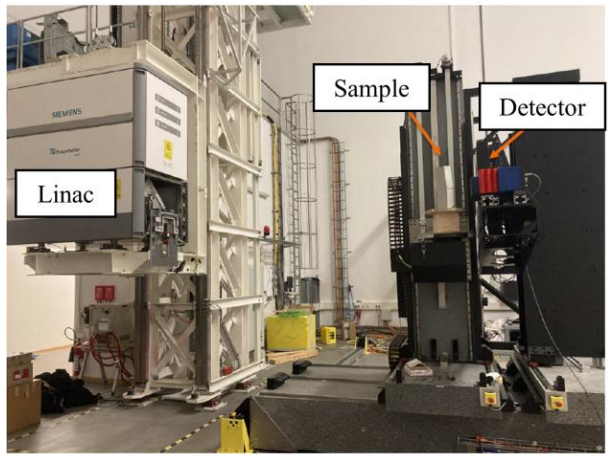

**Fig. 3.** Experimental setup for μCT imaging of the PPFRC sample using the linear accelerator at EZRT in Fürth.

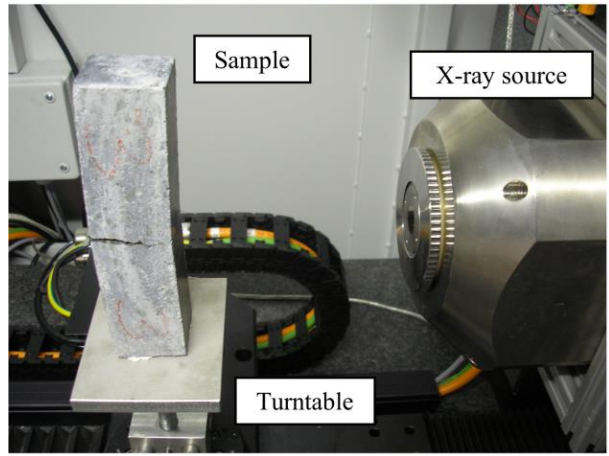

**Fig. 4**. Experimental setup for µCT imaging of the SFRC sample using ITWM's laboratory CT device.

## 2.4. Image and crack characteristics

Table 1 summarizes details about the CT images of the SFRC and the PPFRC samples. We crop both images to remove dark regions in the SFRC image and non-crack regions in the PPFRC image. **Fig. 5** shows one slice of each of the two images. There are no big differences in the characteristics of the cracks. For SFRC, the crack becomes thick (up to 130 voxels corresponding to 3mm at the widest point) and has many thin branches. The side cracks are a desired effect of the steel fibers, which transmit forces into the sample from one side of the crack to the other as described in [33]. The crack propagates from the right to the left side of the specimen, is thick on the right side, thinner at the left edge of the specimen. Detached pieces of concrete are visible in the crack.

As a result of the bending test, the crack in the PPFRC specimen opens widely (at the widest point up to 160 voxels in the reconstructed image corresponding to 9mm, which corresponds to the crack opening at which the laboratory test was stopped). It propagates from top to bottom through the entire sample. There are several branches, most of which are only a few voxels thick. The presence of these secondary cracks proves the influence of the PP fibers on the development of the crack and their high effectiveness as secondary cracks form due to stresses being transmitted by the fibers. In [34] similar observations were described in detail. Finally, within the crack, we observe some loose pieces of concrete.

Another aspect apparent in **Fig. 5** is the difference in gray values of the fiber types as steel and PP absorb X-rays differently. The induced different gray value distributions in the images impact the subsequent analysis of the images significantly.

**Table 1.**
Details for the CT images of the SFRC and the PPFRC samples studied.

| Sample | Gray value range | Size [voxels] | Cropped size [voxels] | Voxel edge length | Original sample size [cm] |
|---|---|---|---|---|---|
| SFRC | 16-bit | 2.046 × 2.046 × 962 | 1.579 × 1.419 × 877 | 25.2μm | 5.16 × 5.16 × 2.42 |
| PPFRC | 8-bit | 2.145 × 2.001 × 2.427 | 1.600 × 1.600 × 1.600 | 60.4μm | 12.91 × 12.05 × 14.61 |

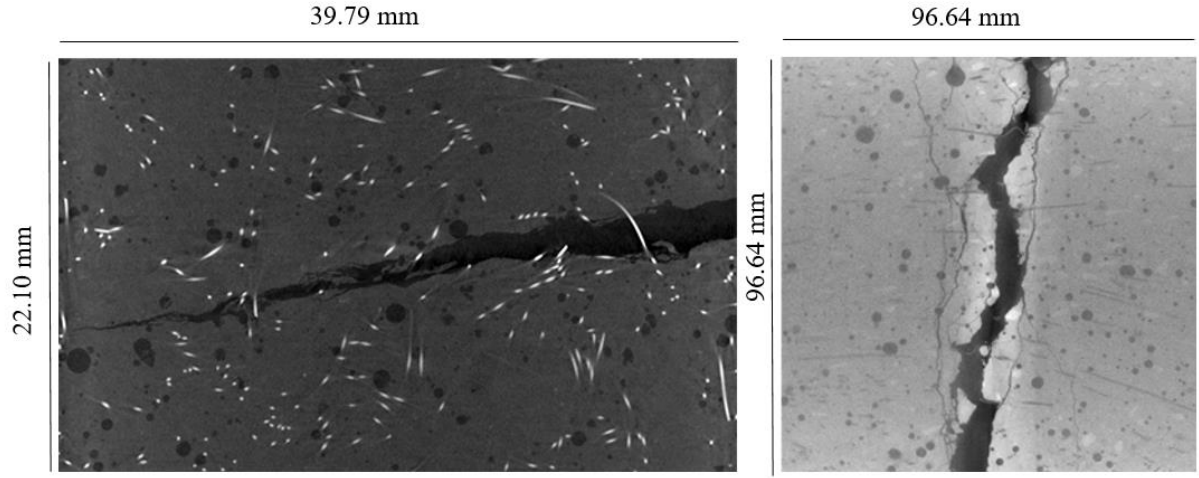

**Fig. 5.** Slices extracted from the investigated CT images (SFRC on the left and PPFRC on the right). Steel fibers appearing very bright due to strong X-ray absorption. PP fibers are dark as PP absorbs X-rays only weakly.

## 3. Semantic segmentation of 3d images using convolutional neural networks

In this section, we provide theoretical background for Deep Neural Networks (DNN) and

Convolutional Neural networks (CNN). We introduce fundamental concepts like semantic segmentation, supervised learning, and training of deep neural networks. Moreover, we describe the method we used for our task of segmentation of cracks in various types of concrete.

### 3.1. Semantic segmentation

The concept image segmentation describes the process of dividing the image into multiple segments [35]. Compared to classification, where the whole image needs to be assigned to a specific class, segmentation involves classification of each pixel of a 2d image or each voxel of a 3d image, as in our case. There are two major types of image segmentation: semantic segmentation, where all objects of the same type are marked using a class label, and instance segmentation, where similar objects get individual labels. Here we deal with binary semantic segmentation of 3d images, where the voxels are classified into one of the two classes background or crack.

### 3.2. Deep neural networks

Neural networks are a special type of machine learning model [36]. They are composed of multiple layers of interconnected nodes, called artificial neurons or units. They are called deep if they have a large number of layers, often several dozens or more. The nodes of the network are assigned weights. The network processes the input through several layers of artificial neurons and each neuron executes a mathematical operation on the input data. The output of the final layer of the network is the so-called prediction. **Fig. 6** shows an example of a neural network architecture.

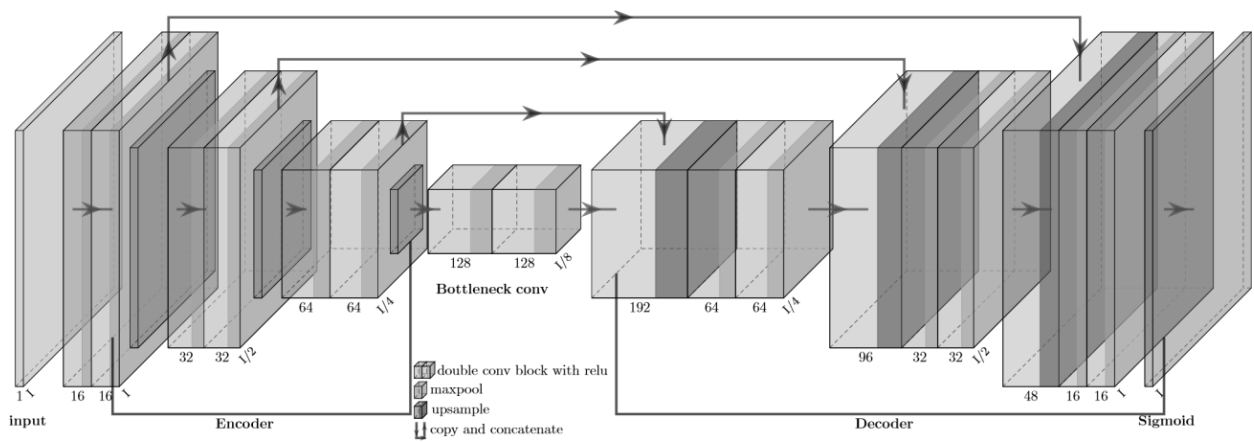

**Fig. 6.** Outline of the 3d U-Net architecture used here.

Each layer of a neural network receives input from the previous layer, processes it, and passes it on to the next layer. The first layer, called the input layer, receives the raw input data. The final layer, called the output layer, produces the final output predictions, in our case a so-called semantic segmentation map – a gray value image of the same dimensions as the input image. For the binary segmentation, the output segmentation map values are between 0 and 1 and represent probabilities of class membership. The layers between input and output layers are called hidden because their inputs and outputs are not directly visible to the outside world.

Between the hidden layers, the activation function decides whether the current output should be activated (i.e., passed on as input to the next layer). The activation function is what gives neural networks the ability to learn and make decisions, as it allows them to determine whether certain input patterns are relevant for the task. There are many activation functions and the proper one needs to be chosen depending on the task. Here, we use Rectified Linear Unit (ReLU) activation, only. ReLU maps any input less than 0 to 0. Input greater than 0 remains unchanged. ReLU is simple to compute and has been shown to be effective in many deep learning tasks because of its non-linearity. The non-linearity of activation functions is one of the reasons for the huge success of deep neural networks, because the modelled phenomena rarely follow the perfect linearity [37]. After the activation function is applied to the layer's outputs, the subsequent batch normalization process normalizes the outputs of the activation function (activations). This is achieved by subtracting the mean and dividing by the standard deviation of the activations. Batch normalization is widely recognized and employed in various applications.

## 3.3. Supervised learning and training of deep neural networks

This subsection summarizes the fundamental ideas of supervised learning and thrives heavily on [38].In supervised learning, the network gets the input data as well as the ground truths, which are the targets for the network and show, what needs to be learned. For our segmentation task, the inputs are gray value CT images and the ground truths are the same images binarized such that the crack is foreground (white) and all other voxels are background (black).

Training a neural network aims at finding the set of weights enabling the best prediction of the output for given input. To achieve this, the network is fed training input. As a rule of thumb, the more data a model has, the better it can learn and generalize. However, there is no minimum amount of data required for a deep learning model. In case of deep learning neural networks, bigger data sets are needed, because complex models adapt to patterns in the training data very well. They might however then perform well only on the seen data, but really poorly on unseen data. This phenomenon is called overfitting. We explain it in more detail at the end of the section.

### 3.3.1. Forward propagation and backpropagation

The input is processed through the network, from the first layer to the last, in a process called forward propagation. Then each layer applies its own set of weights to the input it receives. The output of each neuron is computed by applying an activation function to the dot product of the input and weights.

The weights need to be adjusted and tuned. To this end, there is another process called backpropagation. Backpropagation starts at the last layer and continues to the first layer and calculates the error. The error is the numeric value measured by the loss function comparing the desired output to the weights predicted in the forward propagation process in each layer. This loss function is to be minimized by an optimization algorithm. This optimizer receives the gradient of the loss function with respect to the network weights and updates them such that the loss is reduced.

### 3.3.2. Loss function

There are various types of loss functions. A popular choice for semantic segmentation is the cross-entropy (CE) loss function. In simple terms, this function compares the output probability from the network to the desired probability and increases when the difference between them is high. The CE loss function can be represented as:

$$H(p, q) = -\sum_n p(n) \log q(n) \qquad (2)$$

where *H(p,q)* is the CE between the true probability distribution *p* and the predicted distribution *q*. The probability *p(n)* is the true one of the *n*-th class, while *q(n)* is the predicted probability of the *n*-th class, and log denotes the natural logarithm.

The cross-entropy (CE) loss function is preferable for classification tasks, where the desired outputs are categorical variables (1 if the output belongs to the given category, 0 otherwise), while the maybe at first glance more natural Mean Squared Error (MSE) is better suited for regressions, where the desired outputs are numerical values. The CE is calculated according to Formula (1). Compared to the MSE, CE loss values are higher due to the use of logarithm. This emphasizes the information for the network, that the prediction is incorrect.

### 3.3.3. Optimization algorithms

As for the loss function, there are various optimization functions, one of the simplest being gradient descent as explained e. g. in [39]. More sophisticated are stochastic gradient descent (SGD), Adam, and RMSprop. Each optimization function has its own set of hyperparameters that can be tuned. The learning rate is a common hyperparameter that controls the step size of the weight updates. At a higher learning rate, training finishes faster. However, increasing it too much can cause the model to converge to a local minimum or even diverge as well.

### 3.3.4. Epoch, batch size

As the model is trained, the value of the loss function should decrease, which means that the model is getting better at solving the task. Training is repeated many times in so-called epochs. Within one epoch, the whole network is trained on the complete training dataset. However, the whole training dataset is hardly ever fed to the network at once in the training process. Usually, the data set is divided into batches comprising only a specified, significantly smaller number of training data. Only one batch at a time is applied to the network for training. One epoch continues until the whole dataset, divided into batches, has been applied to the network. The complete training consists of a specified number of epochs.

### 3.3.5. Model performance

Once the model has been trained, it can be tested on new, unseen data to see how well it generalizes to real-world situations. If the network performed well on the training data, but generalizes poorly on the unseen data, the model is overfitted. Overfitting may occur when the model is too complex or if the model has been trained for too many epochs. The opposite phenomenon called underfitting occurs when the model is too simple to capture the complexity of the data, or if the model is not trained for sufficiently many epochs. As a result, the model is not able to capture the underlying relationships in the data, and it will perform poorly on both the training and the test data.

## 3.4. Transfer learning for improving deep neural networks

Transfer learning is a machine learning technique transfering knowledge from an already trained model to a new model, designed to solve a similar kind of problem [40]. In practice it means that the weights from one model are used to speed up the training process of the new model. This can be useful when training a model from scratch would be time-consuming or require a lot of data, or when there is a shortage of labeled training data for the new task. Fine-tuning is just another term for transfer learning. It usually refers to the situation, when the size of the new data set is small and therefore training a model from scratch instead would induce overfitting. Fine-tuning can improve model prediction results. For example, in [41], the performance of a CNN in a medical application was investigated experimentally. One model pre-trained on a general, non-medical image data set and subsequently fine-tuned, the other one trained from scratch on the task-specific data set. The fine-tuned model outperformed the one trained from scratch with the effect getting more and more apparent when the training data was reduced.

## 3.5. Convolutional neural networks

Expanding [42], we delve into the concept of convolutional neural networks and their individual components. Machine learning techniques have recently induced a leap forward in image segmentation. This is mainly due to the CNN - special deep neural networks. CNN are dedicated to work with image data and have been proven to achieve satisfying segmentation results in 2d images for example in [43] and [44]. They were successfully applied to segment crack structures in [45] and [46]. Most state-of-the-art research involving image processing, especially in medical applications, uses CNNs in order to segment the component of interest in the images.

### 3.5.1. Convolutional layer

CNNs feature convolutional layers - a type of neural network layer primarily used in image recognition tasks. They are called convolutional because they convolve the input with a kernel (also called filter or weight matrix) to extract certain features from the image, like specific shapes, edges or elements. More precisely, the kernel is a small matrix of weights, applied to the input image data by sliding it across the width and height (and thickness for 3d) of the input image and computing the dot product of the matrix with the surrounding input image values at each pixel or voxel position, see **Fig. 7**. The values in the kernel matrix are learned during the training process and are used to extract features from the input image. The convolution generates a 2d or 3d output map called the feature map, which is a grid of numbers representing the filtered output of the corresponding position of the input image. The hyperparameter stride controls by how many pixels/voxels the kernel is moved each time it is applied. The stride determines the size of the output feature map. A larger stride results in a smaller feature map, since the kernel is applied less often. A smaller stride induces a larger feature map, because the kernel is applied more often.

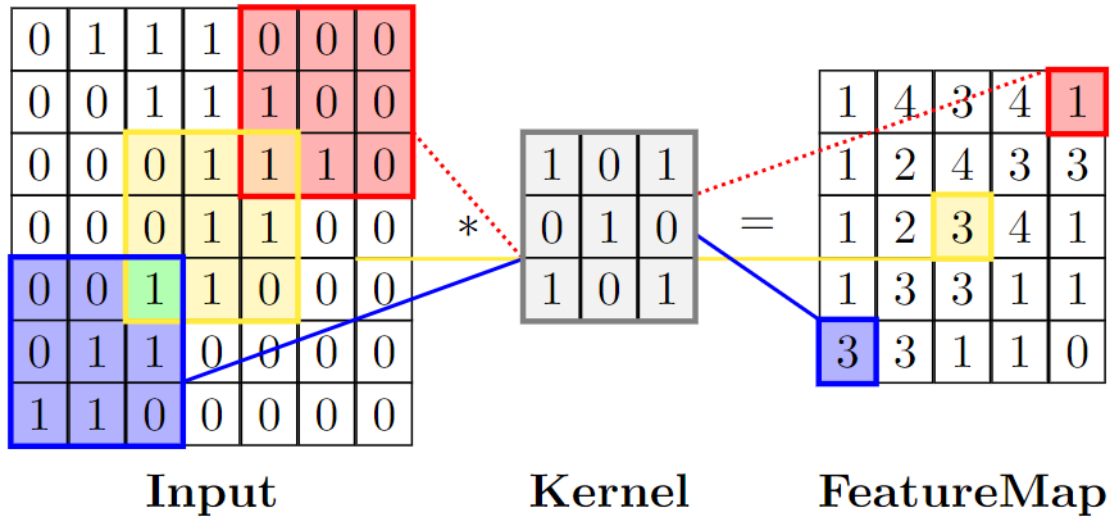

**Fig. 7.** Scheme of a convolution as applied in convolutional layers. The dot products of rectangular areas of input and kernel matrices are stored at the corresponding position in the feature map matrix.

### 3.5.2. Transposed convolution layer

To inflate the input data, a transposed convolution layer may be applied. It inserts zeros between the entries of the input data, which effectively increases the size of the input and allows a kernel to be applied more often. The output feature map produced by a transposed convolution layer is larger than the input, which allows the spatial dimensions of the input data to be expanded.

### 3.5.3. Pooling layer

Besides the eponymous convolutional layers, CNNs feature also pooling layers. A pooling layer in a CNN reduces the spatial dimensions of the feature maps produced by the convolutional layers. The main purpose of pooling is to reduce the computational cost of the network, as well as to make the features detected by the convolutional layers more robust to small variations in the input image, like rotations or shifts. The hyperparameters of pooling layers are again kernel and stride.

However, pooling kernels do not have weights. Thus, there is nothing to be learned in these layers. The kernel is a matrix of a specified shape, applied to the feature map output by the convolutional layer to extract a specified value from the rectangular region of the feature map, highlighted by the kernel matrix. The kernel matrix of the pooling layer slides across each dimension of the feature map, depending on the chosen value of stride parameter. Several types of pooling operations can be applied like max pooling and average pooling. In max pooling, the maximum value in a rectangular or cuboidal region of the feature map is selected and used as the output value for that region. In average pooling, values in the rectangular or cuboidal region of the feature map are averaged.

# 4. A CNN for segmenting concrete cracks in CT scans

In this section we describe the 3d U-Net [17] and how to adapt it for segmentation of concrete cracks. We explain the data creation process. Moreover, we justify the choice of parameters.

## 4.1. Challenges of segmentation of cracks in 3d CT scans

The microstructure of concrete is very heterogeneous, thus CT images of concrete strongly vary locally. The microstructural features, which the CNN learns during the training process, vary widely, too. There can be aggregates, pores, reinforcement, fibers or cracks. Depending on the concrete type, the gray values of those features in the CT images differ a lot. Therefore, a model trained on CT scans of one type of concrete may not perform well on another type of concrete, especially if concrete matrix and structure differ. The gray values of pores and cracks depend on the concrete matrix. CT scans of normal concrete (NC) and high performance concrete (HPC), which were used to train the 3d U-Net in [12] are characterized by a bright structural part and dark cracks and pores. Reinforcing fibers like in SFRC change the gray value distributions significantly compared to images of NC or HPC samples without reinforcing fibers. In the CT scans, the steel fibers appear very bright due to the strong X-ray absorption of steel. As a consequence, both the concrete matrix and the cracks and pores appear much darker and the gray value contrast between these components decreases. Another example of concrete that might cause difficulties for neural networks, is PPFRC. The gray values of the PP fibers are very similar to the gray values of thin cracks in the CT scans. Both fiber reinforced concrete types are shown in **Fig. 8** and **Fig. 9**.

The 3d U-Net trained in [12] can segment the voxels belonging to the crack class very well in NC or HPC. However, it does not cope that well with real cracks in fiber reinforced concretes. Thus, our 3d U-Net is not yet sufficiently robust to be applicable to every type of concrete. **Fig. 8** illustrates the problems occurring: thin crack regions in SFRC are not detected. Moreover, the model assigns the bright fibers and the edges of pores to the crack class. These concrete matrix features were not included in the training data, thus the model might randomly classify it to some class. Here, the network wrongly classifies the fibers as cracks. The model trained on a dataset including the FRC images performs much better, since the network actually knows that fibers were not labelled as crack in the ground truths.

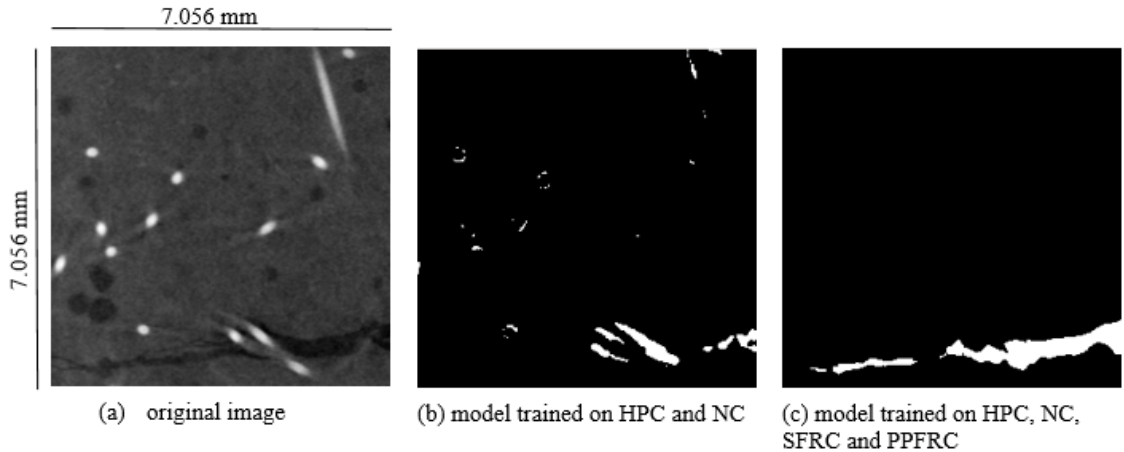

**Fig. 8** 2d slices of 3d images showing segmentation results for the SFRC sample. (b) and (c) predictions of the adapted 3d U-net models trained on 24 images. (b) trained on 12 HPC and 12 NC images. (c) trained on NC, HPC, PPFRC, and SFRC (6 images of each).

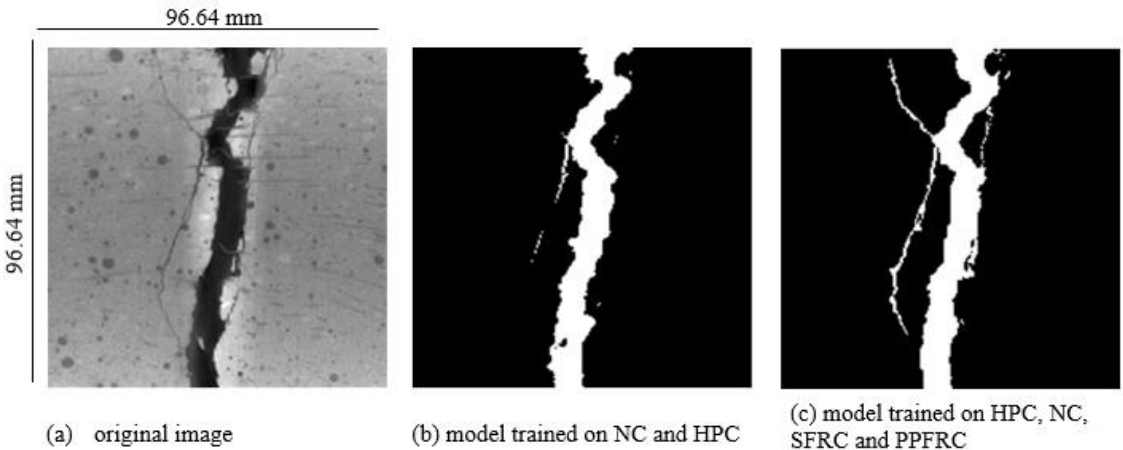

**Fig. 9** 2d slices of 3d images showing segmentation results for the PPFRC sample. (b) and (c) predictions of the adapted 3d U-net models trained on 24 images. (b) trained on 12 HPC and 12 NC images. (c) trained on NC, HPC, PPFRC, and SFRC (6 images of each).

4.2. Training data

In the final dataset, we have 24 spatial images, each $256^3$ voxels large. We use images of NC, HPC, PPFRC, and SFRC to simulate the backgrounds for several crack widths. There are six images for each of these four types of concrete, namely three images containing a single crack of widths one, three, and five voxels, respectively, as well as three images with double cracks, again of widths one, three, and five voxels. Double cracks occur in the same or different planes. **Fig. 10** displays representative slices. In order to reduce the computational load, we divided the images into patches and trained the network on each patch size. To mitigate potential edge effects, we maintained an overlap of 14 voxels between adjacent patches.

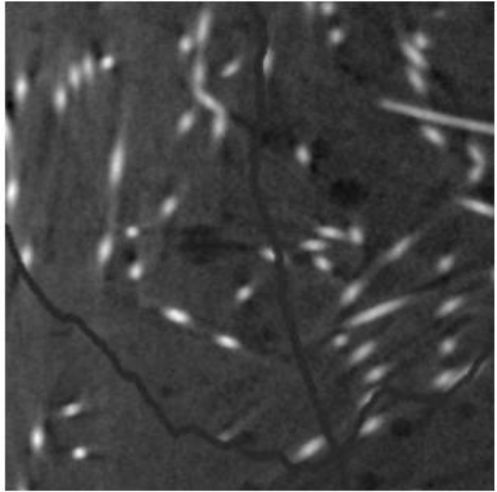
(a) Original image

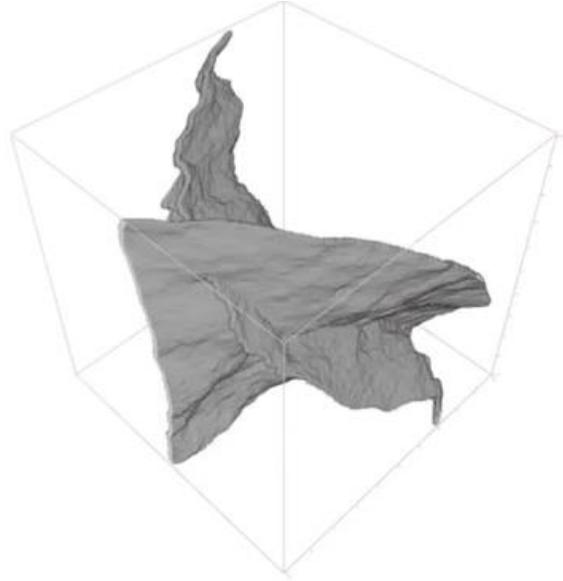
(b) 3d rendering of the crack

**Fig. 10** 2d slice and 3d rendering of 3d semi-synthetic image of size $256^3$ with two cracks used for training.

The models in the previous works [12], [13], [14] were trained on patch size 64, only. This time we investigate the influence of that parameter on the segmentation results. We investigate patch sizes 128, 64, and 32 voxels. The maximum patch size 128 voxels is chosen such that we do not get into memory issues, 64 voxels is the one used before, and 32 voxels seems to be small but still reasonable. The smaller the patch size, the longer is the training process. Because of that, we need to reduce the dataset when training on patch size 32 voxels. To shorten the training time, we remove the patches, which contain fewer voxels belonging to the crack class than 0.005% of all voxels – a measure that had not to be taken when using bigger patch sizes. We do not reduce the patch size further to 16 voxels, as the training time would extend significantly and the dataset would have to be reduced even further.

Table 2 yields details about the size of dataset for each patch size. The results obtained from each model are presented in Section 5.

**Table 2**
The table representing the dataset for each patch size.

| Patch size | No. of patches with crack voxels | No. of patches without crack voxels | No. of patches in the training |
|---|---|---|---|
| 128 | 507 | 69 | 576 |
| 64 | 2.385 | 2.223 | 4.608 |

| 32 | 10.464 | 26.400 | 10.395 |

### 4.3. A dedicated CNN architecture for 3d image segmentation: 3d U-Net

As mentioned above, we used a 3d U-Net CNN architecture for our semantic segmentation task of identifying real cracks in concrete. 3d U-Net is the 3d version of the famous U-Net described in [17]. It has been applied successfully in various fields although originally invented for segmenting medical images. The name "U-Net" is derived from the fact that it includes a contracting path (encoder) that reduces the size of the input image and an extracting path (decoder), which expands the image, giving it the typical U-shape. The sketch of a 3d U-Net architecture can be observed in **Fig. 66**.

We use three convolution blocks in each path, see version of the 3d U-Net architecture in **Fig. 6**. The number of filters is doubled in each convolution block of the contracting path and it is halved in each block of the extracting path. This results in the number of input filters being the same as the number of output filters. Convolutions with cubic 3d kernels of edge length three voxels (short size three) are performed twice in the encoder path, followed by batch normalization and ReLUs. Subsequently, a maximum pooling layer with stride two is applied for downsampling. At the beginning of the expansive path, a 3d transposed convolution is employed with a kernel size of three for upsampling. Afterwards, the reduced feature map obtained from the contracting path is concatenated. Following each stage, two convolutional layers are utilized, both with a kernel size of three. These layers are then followed by batch normalization and ReLU activation. Following the extraction path, an additional convolutional layer with a kernel size of three is utilized. As a consequence, our 3d U-Net has a total of 2.042.689 trainable parameters.

A batch size of two and the Adam optimizer introduced in [47] were employed. The optimizer starts with an initial learning rate of 0.001, and the learning rate is halved every five epochs. We therefore checked the model performance after every five epochs qualitatively visually. This way we found that at least 20 epochs of training are needed. However, five more epochs led to a significant increase in false positives: many fibers got wrongly segmented as cracks.

### 4.4. Fine-tuning

We use fine-tuning to teach our model to detect thick cracks better. More precisely, we use cracks that we segmented in previous works to further train the model from Section 4.3 trained on images with cracks of fixed width only. This fine-tuning or additional training should adjust the model to successfully predict real cracks, especially the thick ones. **Fig. 11** provides the example image used for fine-tuning. The derivation of such semi-synthetic images is described in [20]. Additional 10 epochs suffice since the model has already been trained on the semi synthetic data of fixed width cracks. We keep the same parameters as for basic training.

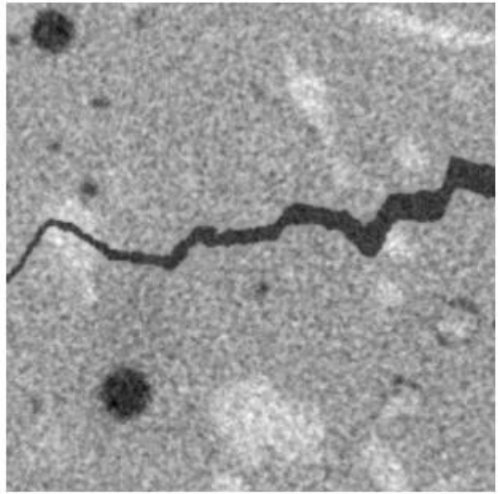

(a) Original image

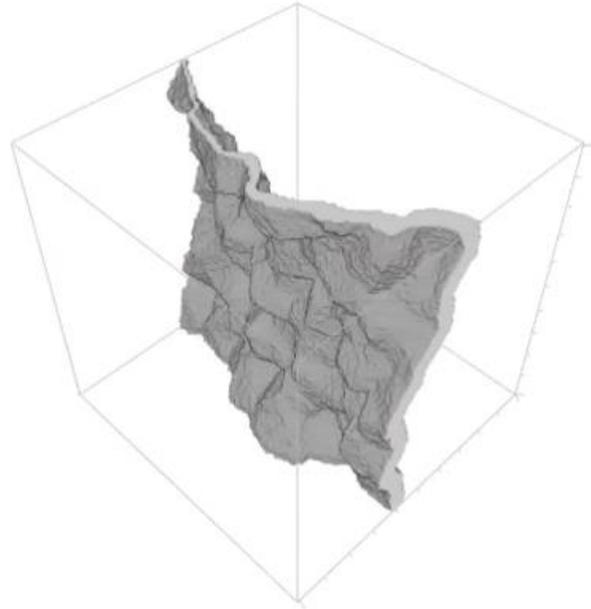

(b) 3d rendering of the crack

**Fig. 11**. 2d slice and 3d rendering of 3d semi-synthetic image of size $256^3$ with multi scale cracks used for fine-tuning.

### 4.5. Hyperparameter adjustment

As explained in Section 2, the real cracks observed in CT data can be considerably wider than the maximum width of five voxels covered by our semi-synthetic training data. To achieve satisfying results in segmentation of real cracks, a multi-scale approach is used. That is, the network is applied on downscaled versions of the original image. We have trained four models: models trained on three patch sizes and a fine-tuned model. For ease of reading, we name the models trained on patch sizes 32, 64, and 128 voxels for short M32, M64, and M128, respectively, and the fine-tuned model MF.

We run M32, M64, and M128 on the downscaled versions of the original SFRC and PPFRC images for scales in {0.0375, 0.0625, 0.125, 0.1875, 0.25}. This results in 5 images of different sizes. We restore them to the original image size by spline interpolation. We use the voxelwise maximum of the prediction images with float voxel values in [0,1] to derive the final prediction and apply a threshold of 0.5, to get the binary image. Voxels with gray values below 0.5 are considered to be background, voxels with gray values above 0.5 are assigned to the crack class. At this point, we have two binary output segmentation maps of the SFRC and PPFRC samples. We extract the largest connected component from each of two images, to get rid of unnecessary noise and obtain a smoother prediction.

For the fine-tuned model, we do not need that much downscaling. In this case we run the model on downsampled images for scales in {0.075, 0.125, 0.25, 0.375, 0.5} and we repeat the steps described before. For the PPFRC sample, we crop out the boundary regions, since many false positives were detected in that area due to the artifacts from CT imaging.

## 5. Results

In this section we describe the segmentation results qualitatively, based on visual impression from 2d slices as well as 3d renderings. The 3d renderings carry information about the overall model performance and provide the big picture of the segmented crack. In the 2d slices, we can inspect some elements better, for example thin cracks. We do not have ground truths for the original images since it is extremely time consuming and error prone to annotate the crack voxels manually. Hence, we cannot assess the models numerically. We can nevertheless compare the results of four models, namely the three models trained on patch sizes 32, 64, 128 voxels as well as the model pre-trained on patch size 64 voxels and later fine-tuned as described in Section 4.4.

### 5.1. Polypropylene fiber reinforced sample

**Fig. 12** contains 2d slices of segmentation results for the PPFRC image, while 3d renderings are shown in **Fig. 13**.

M32's segmentation maps look in general good. The connectivity of the segmented cracks (thin ones as well as the thick, main crack) seems to be preserved. The thin crack disappears only in some areas, where it is barely visible in the original image. In the 3d rendering, we notice however some fibers misclassified as cracks.In the 2d slice view, segmentation maps derived by M64 do not differ much from those of M32. Thin cracks have been segmented as well as the main crack. However, the 3d renderings reveal that the crack opening at the very bottom is lacking a small central part. On the other hand, fewer fibers have been misclassified.

M128 is generally underfitted as **Fig. 12** shows. Even though the main crack has been segmented, the thin cracks surrounding it are segmented very poorly. **Fig. 13** reveals that the M128 segmentation map lacks many important elements.

For the fine-tuned model MF, the 2d slices feature some pores and many fibers misclassified as crack. Even though main crack and thin cracks have been segmented correctly, the model seems to be overfitted. The 3d rendering confirms this impression.

To sum up, M32 and M64 perform the best for the PPFRC sample. M128 is underfitted, while MF is overfitted.

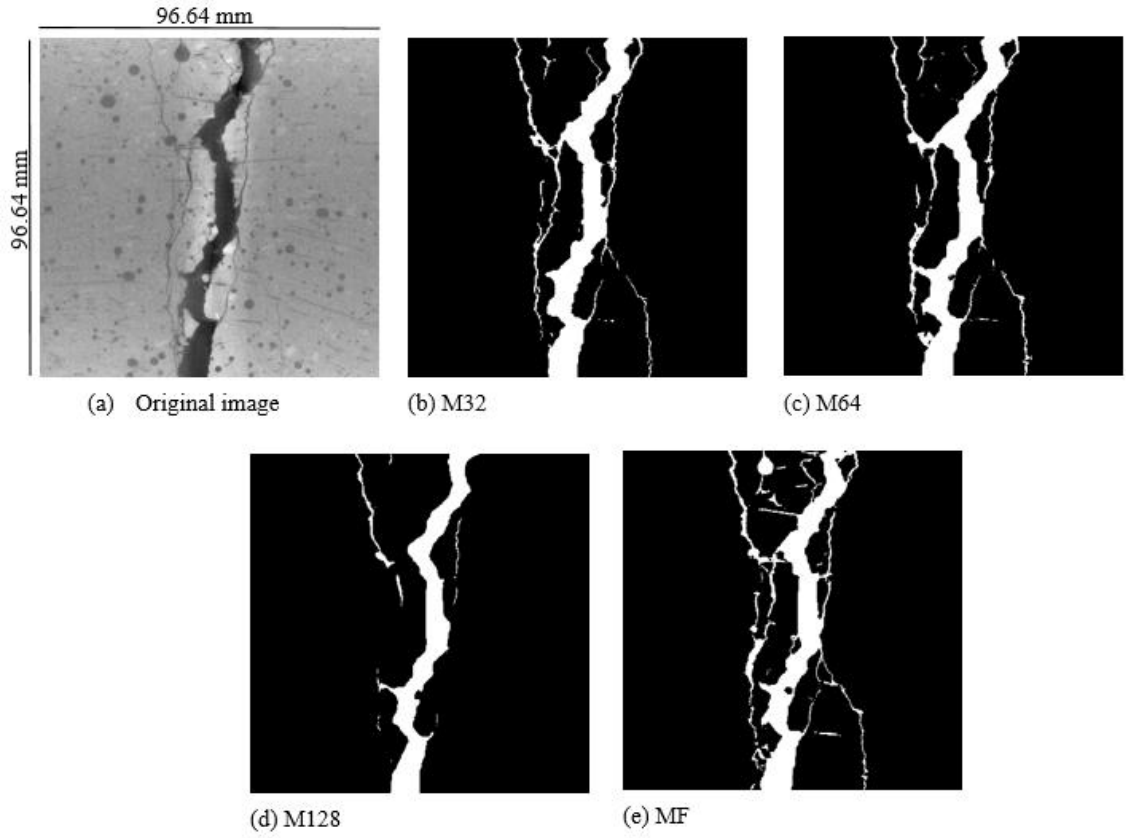

**Fig. 12.** 2d slices of 3d segmentation results of the PPFRC sample obtained by models trained on data with fibers.

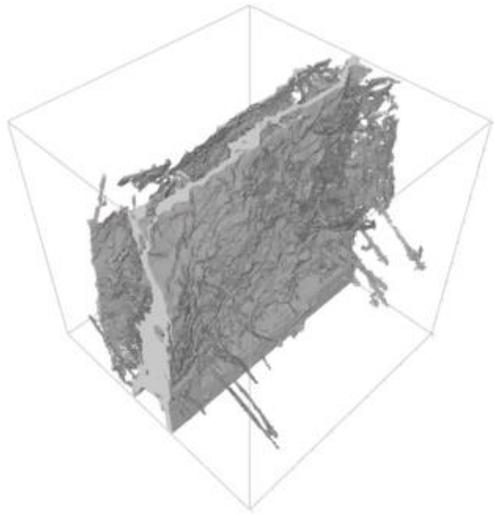

(a) M32

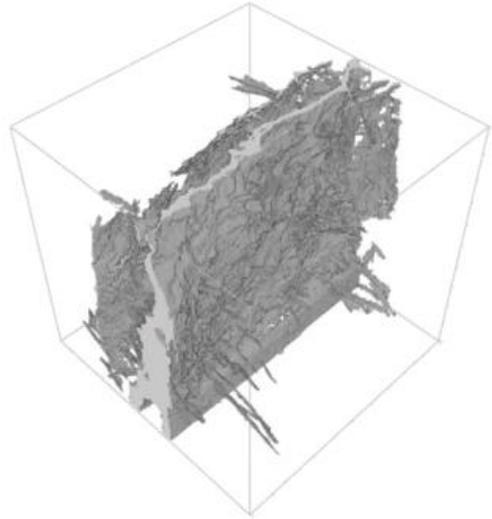

(b) M64

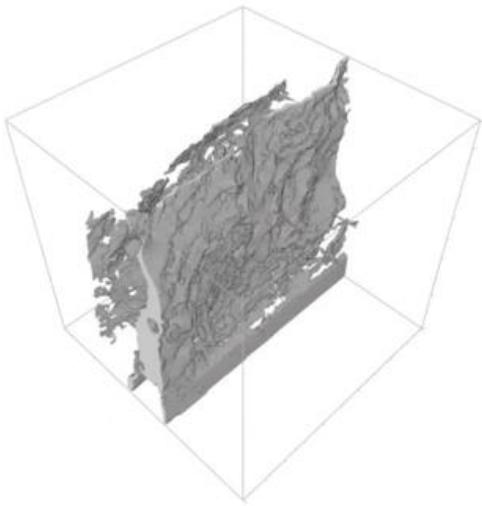

(c) M128

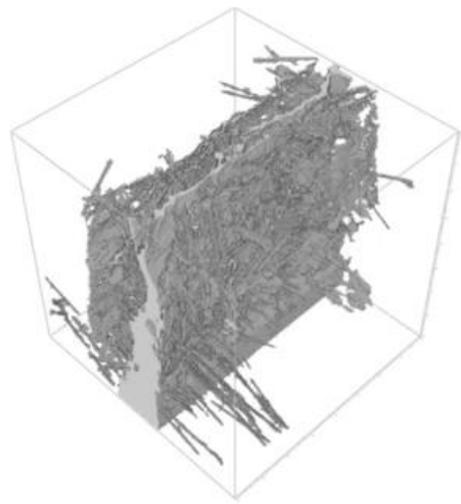

(d) MF

**Fig. 13.** 2d slices of 3d segmentation results of the PPFRC sample obtained by models trained on data with fibers.

## 5.2. Steel fiber reinforced sample

2d slices and 3d renderings of segmentation results for the SFRC sample can be observed in **Fig. 14** and **Fig. 15**, respectively.

M32 did not segment the main crack at all, only the thin part of it. The overall performance of M64 is very satisfactory. The crack is segmented along its whole length and misclassifications of fibers or pores occur barely. M128's segmentation maps are not precise enough. The 3d rendering reveals holes in the segmented crack. The crack segmented by MF is continuous and segmented along its whole length. Overall this model performs as good as M64.

Summing up, for the SFRC sample, the best results were obtained by M64 and MF as well. The models M32 and M128 fail at segmenting the thick part of the crack. Only very few steel fibers or pores were misclassified by all four models.

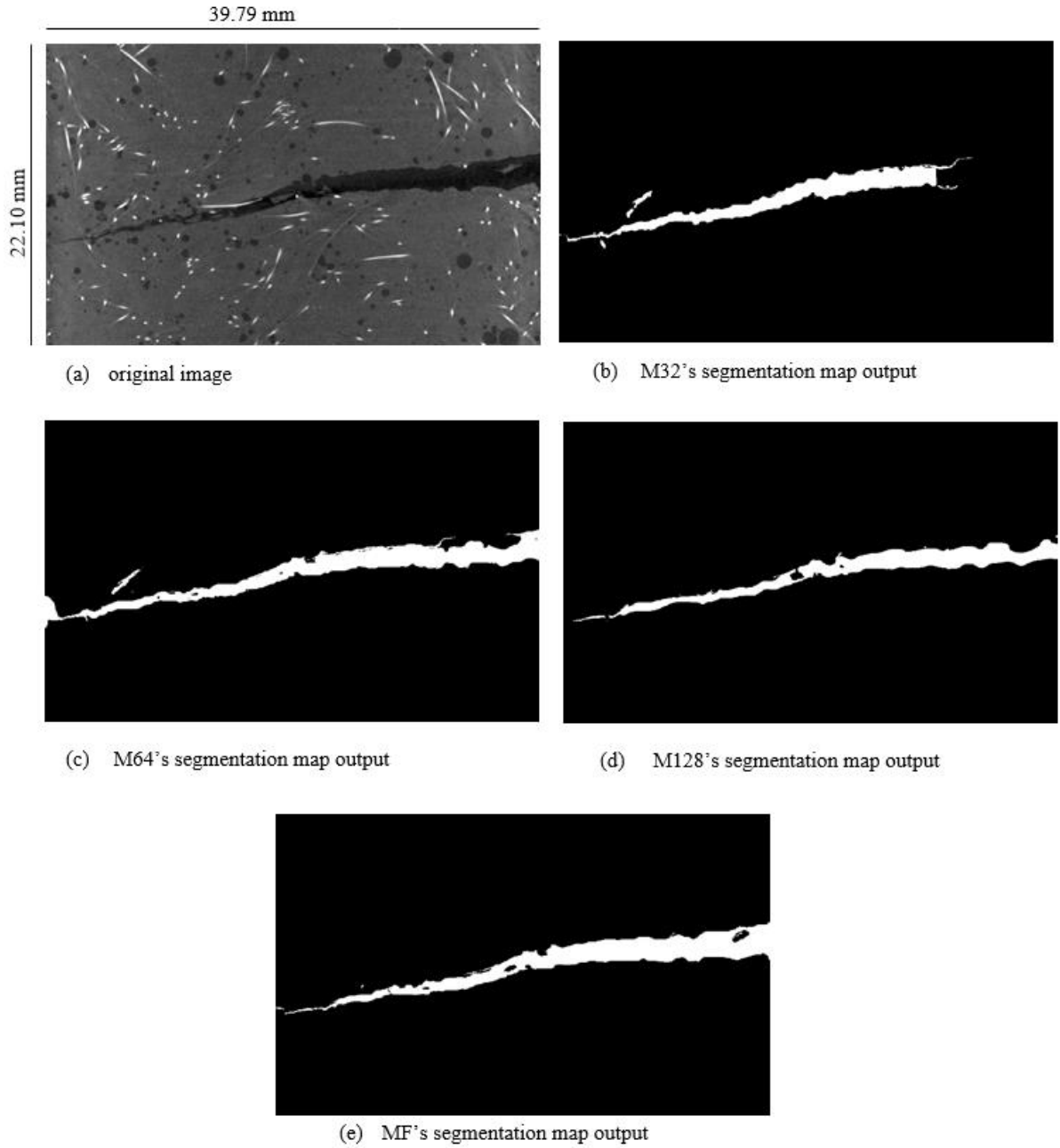

**Fig. 14.** 2d slices of 3d segmentation results of the SFRC sample obtained by models trained on data with fibers.

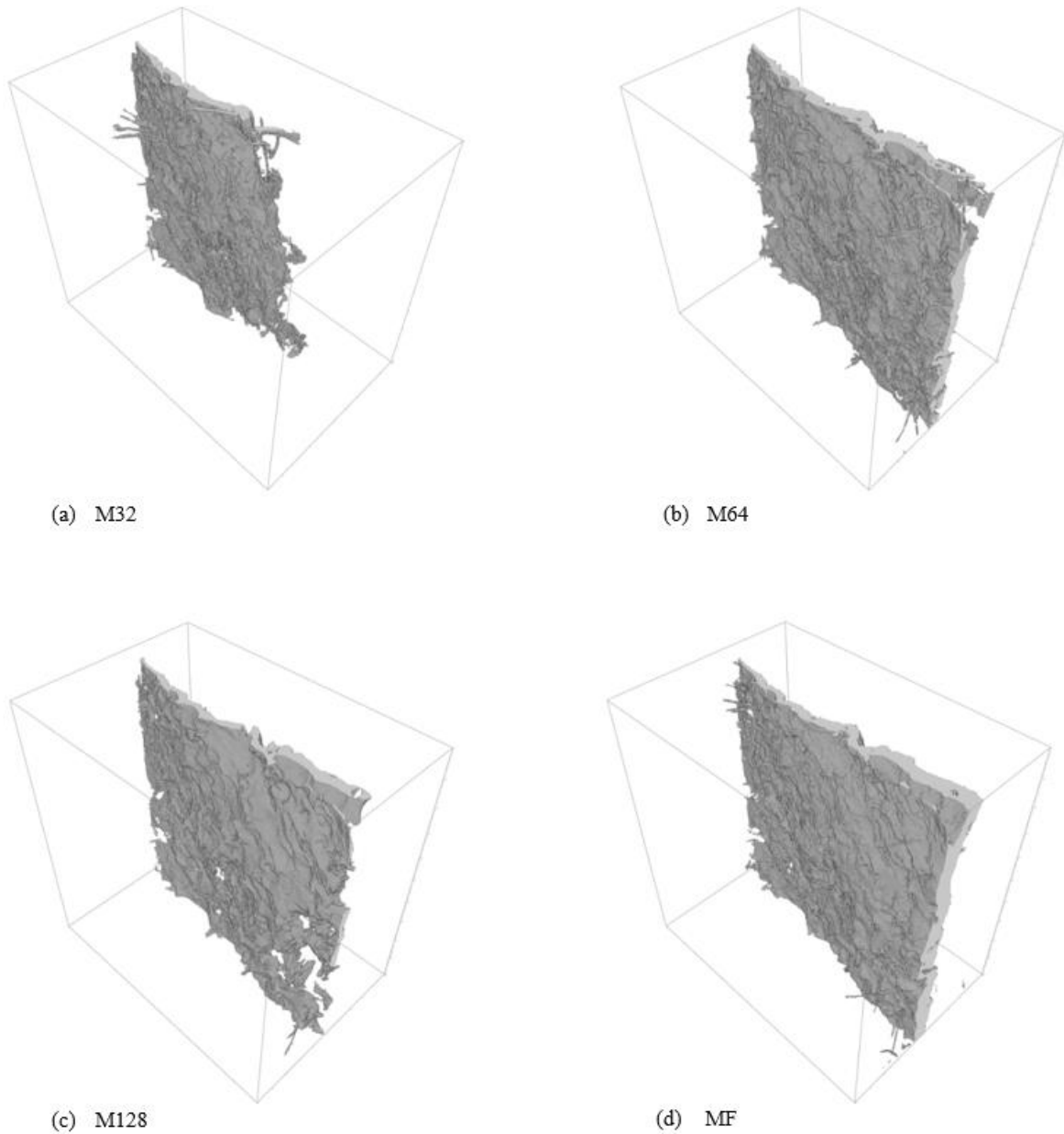

**Fig. 15.** 2d slices of 3d segmentation results of the SFRC sample obtained by models trained on data with fibers.

## 6. Conclusions

We tested the convolutional neural network 3d U-Net as a tool for semantic segmentation of crack structures in 3d images of samples of two types of fiber reinforced concrete – steel fiber and polypropylene fiber reinforced concrete. Both samples feature cracks induced by laboratory tests, but the concrete matrix types and the types of fibers differ. In [12], [13], and [14] the segmentation methods were applied on concretes without fibers, only. Here, we proved that 3d U-Net trained on semi-synthetic data also works for concretes with fibers, given it has seen FRC samples during training, too.

Moreover, we explored the influence of patch size on training process and time as well as on the results. A patch size of 64 voxels turned out to yield the most reliable and stable results.

The results presented here open the opportunity to analyze the crack opening in the internal structure of the concrete truly spatially and non-destructively. So far, this has only been possible on the surface of the sample. The crack opening can now be determined with a high accuracy at any point in the sample, even if the crack has a number of branches. Analysis is subject of further research. In particular, in-situ tests will be conducted to determine the fracture energy.

## Acknowledgments

We would like to thank Michael Salomon of Fraunhofer EZRT in Fürth and Franz Schreiber of Fraunhofer ITWM in Kaiserslautern for CT imaging and the staff of the Civil Engineering Department of University in Kaiserslautern-Landau for laboratory assistance.

## Conflicts of interest

The authors declare no conflict of interest.

## Data availability

No public dataset was used in this research. The data belongs to the University of Kaiserslautern-Landau and Department of Image Processing, Fraunhofer Institute for Industrial Mathematics.

## Funding

This work was supported by the German Federal Ministry of Education and Research (BMBF) [grant number 05M2020 (DAnoBi)].